\documentclass[runningheads]{llncs}

\usepackage{eccv}

\usepackage{eccvabbrv}

\usepackage{graphicx}
\usepackage{booktabs}

\usepackage{multirow}
\usepackage{amsmath}
\usepackage{pifont}
\usepackage{algorithm}
\usepackage{algpseudocode}

\usepackage[accsupp]{axessibility}  

\usepackage[pagebackref,breaklinks,colorlinks,citecolor=eccvblue]{hyperref}

\usepackage{orcidlink}
\usepackage{enumitem}

\newcommand{\cmark}{\ding{51}}%

\newcommand{\model}{\textsc{VidAssist}\xspace}
\newcommand{\baseIOshort}{LLM}
\newcommand{\basePlanshort}{LLM Agent}
\newcommand{\baseIO}{LLM Baseline}
\newcommand{\basePlan}{LLM Agent Baseline}

\begin{document}

\title{Propose, Assess, Search: Harnessing LLMs for Goal-Oriented Planning in Instructional Videos}

\titlerunning{\model}


\author{Md Mohaiminul Islam$^{1,2}\thanks{Work done during an internship at Meta.}$\orcidlink{0000-0003-3337-0000}
\quad
Tushar Nagarajan$^1$\orcidlink{0000-0002-1627-3842}
\quad
Huiyu Wang$^1$
\\
Fu-Jen Chu$^1$\orcidlink{0000-0002-3290-8094}
\quad
Kris Kitani$^1$\orcidlink{0000-0002-9389-4060}
\quad
Gedas Bertasius$^2$\orcidlink{0000-0003-1800-4790}
\quad
Xitong Yang$^1$\orcidlink{0000-0003-4372-241X}
}

\institute{
$^1$FAIR, Meta \quad\quad $^2$UNC Chapel Hill
\\
\href{https://sites.google.com/view/vidassist}{https://sites.google.com/view/vidassist}
}

\authorrunning{M. Islam et al.}


\maketitle

\begin{abstract}
    Goal-oriented planning, or anticipating a series of actions that transition an agent from its current state to a predefined objective, is crucial for developing intelligent assistants aiding users in daily procedural tasks. The problem presents significant challenges due to the need for comprehensive knowledge of temporal and hierarchical task structures, as well as strong capabilities in reasoning and planning. To achieve this, prior work typically relies on extensive training on the target dataset, which often results in significant dataset bias and a lack of generalization to unseen tasks. In this work, we introduce \model, an integrated framework designed for zero/few-shot goal-oriented planning in instructional videos. \model leverages large language models (LLMs) as both the knowledge base and the assessment tool for generating and evaluating action plans, thus overcoming the challenges of acquiring procedural knowledge from small-scale, low-diversity datasets. Moreover, \model employs a breadth-first search algorithm for optimal plan generation, in which a composite of value functions designed for goal-oriented planning is utilized to assess the predicted actions at each step. Extensive experiments demonstrate that \model\ offers a unified framework for different goal-oriented planning setups, e.g., visual planning for assistance (VPA) and procedural planning (PP), and achieves remarkable performance in zero-shot and few-shot setups. Specifically, our few-shot model outperforms the prior fully supervised state-of-the-art method by +7.7\% in VPA and +4.81\% PP task on the COIN dataset while predicting 4 future actions. Code, and models are publicly available at \url{https://sites.google.com/view/vidassist}. 
\end{abstract}
\section{Introduction}
\label{sec:intro}

\begin{figure}[t]
  \centering
  \includegraphics[width=1\textwidth]{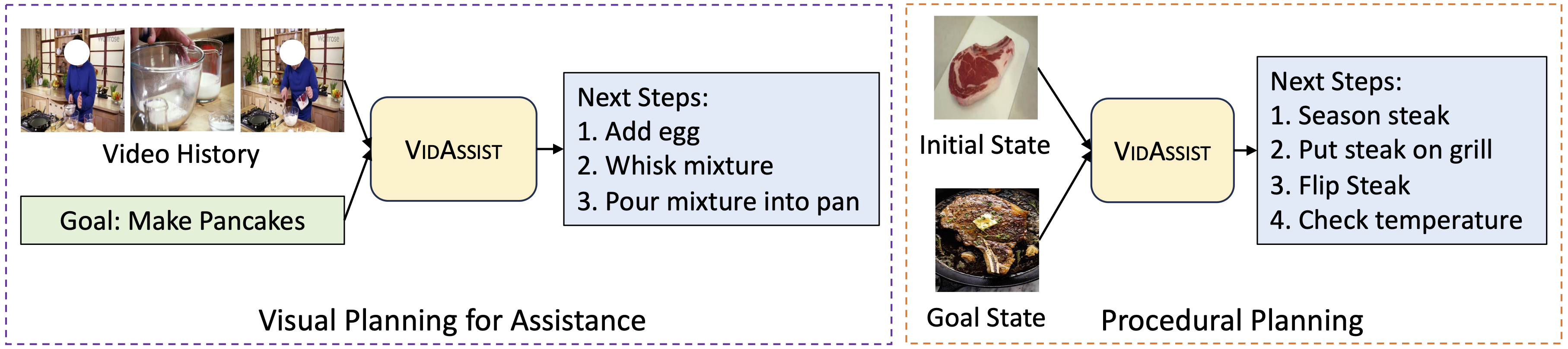}
  \caption{Goal-oriented planning aims to generate action plans to achieve a given goal based on the visual observations. The task unifies two setups that have been \textit{independently} explored in prior literature: Visual Planning for Assistance~\cite{patel2023pretrained} (left) and Procedural Planning~\cite{chang2020procedure, bi2021procedure, sun2022plate, zhao2022p3iv, wang2023event, liu2023language} (right).}
  \vspace{-5mm}
  \label{fig:teaser}
\end{figure}

This work addresses goal-oriented planning, a task that requires forecasting action plans to achieve a given goal based on the current observations. Depending on the specifications of the current observation and the desired goal, the task can be instantiated in two different setups, both of which have been \textit{independently} explored in prior literature: (a) Visual Planning for Assistance (VPA)~\cite{patel2023pretrained}, where the current observation involves an untrimmed video history of the user's progress and the goal is described in natural language; and (b) Procedural Planning (PP)~\cite{chang2020procedure}, where the initial and goal states are presented with images (\Cref{fig:teaser}). Both setups are of great interest in developing intelligent embodied systems capable of understanding human decision-making processes and assisting humans in solving procedural tasks, such as cooking or assembling furniture.

Goal-oriented planning can be viewed as a sequential decision-making process\cite{bellman1957markovian}, in which a goal-conditioned policy sequentially predicts the next action based on the observed states. However, learning such a policy from video data poses significant challenges, as it requires a comprehensive understanding of procedural tasks, including the temporal and hierarchical interconnections between action steps. 
To achieve this, prior work typically relies on extensive training on target datasets, utilizing full supervision from action labels, visual representations~\cite{chang2020procedure, bi2021procedure, sun2022plate}, or text description~\cite{zhao2022p3iv, wang2023event, liu2023language}. These methods share a common limitation -- procedural knowledge, when derived from strong supervision, exhibits a significant bias towards the training dataset, particularly when the scale of annotation and the diversity of tasks are limited. This hinders the model's ability to generalize across new data distribution (\eg, action plans of varying granularity) and to adapt to unseen tasks. Note that the capability for zero/few-shot learning in a planning system is essential for its deployment within intelligent embodied assistants or robots~\cite{finn2017one, sunderhauf2018limits}.

In this paper, we propose \model, an integrated framework that tackles goal-oriented planning in a zero- or few-shot learning scenario. Drawing inspiration from human decision-making process~\cite{sloman1996empirical, kahneman2011thinking} and research in robot planning~\cite{huang2022language,ahn2022can}, we design a search-based framework that facilitates deliberate decision-making at each prediction step and ultimately generates the optimal action plan based on search heuristics. Furthermore, our framework leverages pretrained large language models (LLMs) for generating and evaluating action plans, serving as the knowledge base and the tool for assessment. We hypothesize that LLMs, pretrained with a wealth of Internet-sourced text including web knowledge bases (\eg, wikiHow~\cite{wikiHow}), inherently possess extensive knowledge about a variety of procedural tasks and robust reasoning capabilities.

The \model model first processes visual observations and goals through a Socratic approach~\cite{zeng2022socratic, zhao2023antgpt}, \ie, by converting visual inputs into textual descriptions using video/image understanding models. Afterward, we predict the optimal action plan consisting of the future action steps by three key processes. (i) \textit{Propose} -- prompting LLMs to predict $K$ probable subsequent actions in free-form language, conditioned on the information grounded from visual observations/goals as well as the action plan predicted in prior steps. This allows us to leverage the procedural knowledge embedded within LLMs while also accounting for the inherent uncertainty of procedural tasks. (ii) \textit{Assess} -- evaluating each proposed action through a mixture of value functions specifically designed for goal-oriented planning. The value functions assess the coherence and viability of the trajectory of predicted actions utilizing an LLM. They also evaluate the confidence of language generation and its alignment with admissible actions. (iii) \textit{Search} -- identifying the optimal action plan based on the assessment scores employing a breadth-first search (BFS) algorithm. Low-scoring action proposals are pruned dynamically to improve the efficiency of the search algorithm from a potentially very large search tree.

The proposed \model framework offers enhanced planning capabilities than the standard LLM-prompt-based baseline, particularly for long-horizon planning. On the VPA task, our zero-shot model outperforms the LLM baseline by 12.9\% and 6.6\% success rate on COIN and  CrossTask datasets while predicting 3 future actions (i.e., planning horizon 3). Similarly, on the PP task, zero-shot \model\ achieves 7.41\% and 6.62\% higher success rates on COIN and CrossTask datasets for a planning horizon 4. Finally, our few-shot model achieves state-of-the-art performance on the COIN dataset, surpassing the prior best method by 7.7\% in VPA and 4.81\% in PP for a planning horizon 4.
\section{Related Works}
\label{sec:related}

\noindent\textbf{Planning in instructional videos.}
The task of procedural planning (PP) in instructional videos was first introduced in~\cite{chang2020procedure}, where given the start and the end goal image of a task, the model needs to output intermediate steps to achieve the goal. Since then, many follow-up works have been proposed for the procedural planning task by leveraging supervision from action labels and visual features of the intermediate steps~\cite{sun2022plate,bi2021procedure}. However, such supervision requires expensive annotation efforts to localize the intermediate steps within a video, which is challenging to extend to large-scale datasets. Recent work instead relies on language representation for supervision~\cite{zhao2022p3iv,liu2023language}. Moreover, recent work~\cite{patel2023pretrained} has introduced another setting for goal-oriented planning in instruction videos, called visual planning for assistance (VPA), where given the goal and video history of the completed actions, the model needs to output the future steps to complete the task. They proposed a fully supervised language model-based approach for the VPA task. Most prior works in this area developed specialized, fully supervised architectures for different goal-oriented planning tasks (e.g., PP and VPA tasks) in instructional videos. On the contrary, we propose a unified model for various goal-oriented planning tasks (e.g., PP and VPA tasks), which shows strong performance with limited annotated data.

\vspace{2mm}
\noindent\textbf{LLMs for task planning.} Socratic  Models~\cite{zeng2022socratic} pioneered the use of LLMs for multimodal reasoning in a zero-shot/few-shot manner. Inspired by this work, several subsequent works~\cite{zhao2023antgpt, chowdhery2023palm} have been proposed for future action anticipation and forecasting from videos. These methods follow a common high-level approach - translating videos into text descriptions using various video understanding models (action recognition, captioning, etc.) and utilizing LLMs to forecast future actions from the translated text descriptions. Furthermore, utilizing language as an intermediate representation allows efficient processing for long-range videos~\cite{islam2024video, kahatapitiya2024language, zhang2023simple}. However, these methods do not incorporate any deliberate planning mechanism specifically tailored for procedural videos. On the other hand, several works \cite{huang2022language, yao2024tree, song2023llm} in the robotics domain have explored generating action plans for robots and general problem-solving utilizing deliberate planning mechanisms. Drawing motivation from both streams of research, we have developed a unified framework for goal-oriented planning in instructional videos. Our model leverages a Socratic approach for visual understanding and incorporates a deliberate planning mechanism using the introduced propose, assess, and search techniques.

\section{Methodology}
\label{sec:method}

In this section, we provide a detailed description of our framework for goal-oriented planning in instructional videos. We first present a baseline method in Section~\ref{sec:baseline}, which leverages LLMs to generate action plans. The baseline demonstrates promising results in anticipating short-horizon actions, but its performance significantly diminishes for planning over a long horizon. In Section~\ref{sec:model}, we introduce our framework, \model, which combines LLMs with a search-based algorithm to achieve superior results in long-horizon planning.

\subsection{Task formulation}
\label{sec:task}
Given the current observation $\mathcal{O}$ and the goal $\mathcal{G}$, \textit{goal-oriented planning} requires the model to generate an action plan, \ie, a sequence of actions $\mathcal{A} = \left \{a_1, \dots, a_T \right\}$, that transforms the current state to the goal state, where $T$ is the planning horizon. Here, each action $a_t\in\mathbb{R}^C$ is a categorical label with corresponding text description (\eg, ``break eggs"), and $C$ indicates the total number of admissible actions.
Moreover, we focus on zero-shot and few-shot setups since they are more practical in embodied AI applications~\cite{finn2017one, sunderhauf2018limits}. Only action supervision $\mathcal{A}$ is available during training, and it is limited to $N$ samples per task ($N=0$ and $10$ for zero-shot and few-shot learning, respectively).

In real-world scenarios, both $\mathcal{O}$ and $\mathcal{G}$ can take various forms, including video, image, or text. For example, a user can describe their desired goal in language (\eg, ``I want to make pancakes"), or simply show an image describing the goal state (\eg, and image of cooked steak, as shown in Figure~\ref{fig:teaser}). In this paper, we consider the following two types of goal-oriented planning.

\vspace{-5mm}
\subsubsection{Visual Planning for Assistance (VPA)}~\cite{patel2023pretrained}. In this setup, the observation $\mathcal{O}$ is presented as an untrimmed video history capturing the user's progress: $\mathcal{O} = \left\{v_{-H}, \dots, v_0 \right\}$, where $H$ denotes the number of clips. The user's goal is specified through natural language description $\mathcal{G} = \left\{d\right\}$.

\vspace{-5mm}
\subsubsection{Procedural Planning (PP)}~\cite{chang2020procedure, bi2021procedure, sun2022plate, zhao2022p3iv, wang2023event, liu2023language}. In this setup, both the initial and goal state is specified with an image: $\mathcal{O} = \left\{v_0\right\}$, $\mathcal{G} =\left\{ v_T\right\}$.

\subsection{Baseline: LLM-based goal-oriented planning}
\label{sec:baseline}
Prior research has shown that large language models (LLMs), pre-trained on colossal amounts of Internet-sourced data, inherently possess extensive world knowledge which enables them to engage in problem solving and reasoning without the need for additional training~\cite{huang2022language, ahn2022can, yao2022react, zeng2022socratic}. To investigate how LLMs can be applied for goal-oriented planning in instructional videos, we develop a baseline system that first converts visual observations $\mathcal{O}$ and goals $\mathcal{G}$ into text descriptions and then prompts LLMs to generate action plans accordingly.

\vspace{-4mm}
\subsubsection{Understanding visual observations and goals.} We consider the Socratic model~\cite{zeng2022socratic}, a predominant approach for visual understanding and grounding, to represent visual observations and goals for LLM-based planning. The main idea is to convert non-textual modalities (\eg, images, videos) into text for downstream LLMs. Specifically, for visual observations presented as video history, we follow the prior work~\cite{patel2023pretrained} to divide the raw video into fixed-length window clips~\cite{islam2024video} and predict the action category of each clip using an action recognition model $F_{\text{video}}$~\cite{xu2021videoclip}. If consecutive clips have the same predicted action category, we consolidate them into one prediction. Consequently, we create an action history sequence $\tilde{\mathcal{O}} = F_{\text{video}}(\left\{v_{-H}, \dots, v_0 \right\}) = \left\{\tilde{a}_{-H}, \dots, \tilde{a}_0 \right\}$ comprising textual descriptions of predicted actions.
For visual observations/goals that are specified with images, we use an image-based step classifier $F_{\text{image}}$~\cite{liu2023language} to predict the initial state $\tilde{\mathcal{O}} = F_{\text{image}}(\left\{v_0\right\}) = \left\{\tilde{a}_0\right\}$ and the goal state $\tilde{\mathcal{G}} = F_{\text{image}}(\left\{v_T\right\}) = \left\{\tilde{a}_T\right\}$, represented in textual description. Note that in VPA, the desired goal is already provided through a language description, thus eliminating the need for visual understanding: $\tilde{\mathcal{G}} = \mathcal{G} = \left\{ d\right\}$.

\vspace{-4mm}
\subsubsection{LLM-based next-action prediction.} Inspired by prior work on LLM-based task planning for embodied AI~\cite{huang2022language, ahn2022can}, we prompt a state-of-the-art LLM (\eg, Llama-2~\cite{touvron2023llama}) to predict the most probable subsequent action, taking advantage of the procedural knowledge embedded within the pretrained model. We employ Llama-2 since it is the best open-source LLM for diverse tasks; however, our framework can easily incorporate other LLMs. Rather than generating the entire action plan in one run, we employ an autoregressive approach to generate subsequent actions step-by-step until the given goal is achieved. As shown in Figure~\ref{fig:prompt}, our LLM prompt includes (i) the initial observation $\tilde{\mathcal{O}}$, (ii) the goal state $\tilde{\mathcal{G}}$, and (iii) the predicted action plan up to the current step $t$: $\hat{\mathcal{A}}_t = \left\{\hat{a}_1, \dots, \hat{a}_t \right\}$, all in text format. In the few-shot setup, we augment the prompt by appending a set of in-context examples containing goals and their corresponding action plans.

One challenge arises when naively employing LLM-based predictions as the action plan: the output space of LLM is unconstrained, and the prediction expressed in free-form language often cannot be translated to admissible actions. This leads to suboptimal action plans, especially for predicting longer horizon action steps. To address this issue, similar to the previous studies~\cite{huang2022language, liu2023language}, we map the free-form LLM output to the most semantically similar admissible action using an off-the-shelf text embedding model (\eg, Sentence-BERT~\cite{reimers2019sentence}). The admissible action that exhibits the highest cosine similarity with the LLM output will then be included in the predicted action plan.

\subsection{\model}
\label{sec:model}

\begin{figure}[t]
  \centering
  \includegraphics[width=1\textwidth]{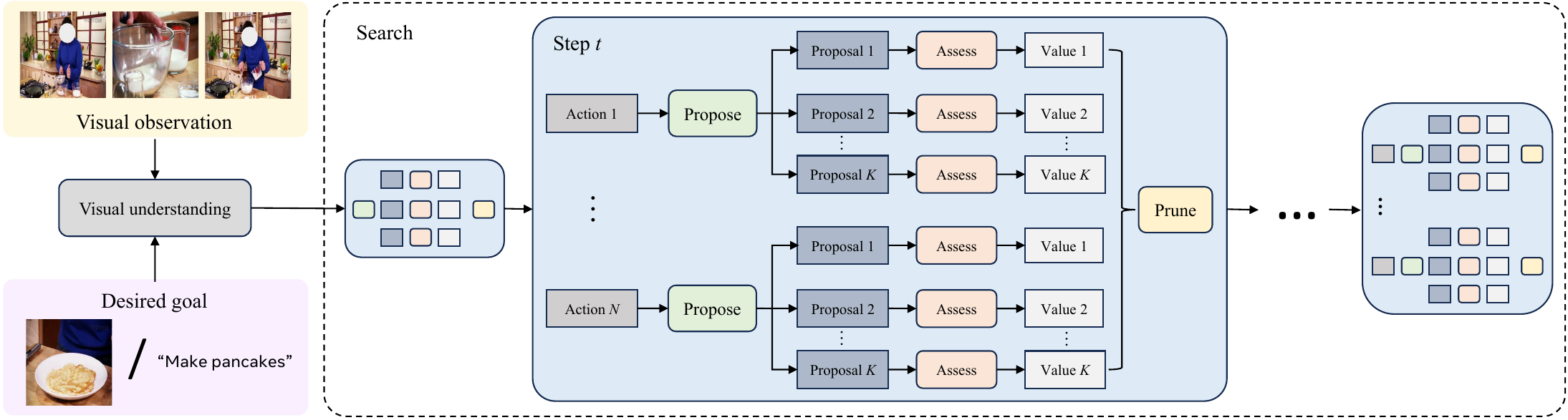}
  \caption{Overview of the \model\ framework. We first process visual observations and goals by transforming the visual inputs into textual descriptions using visual understanding models. Then, we leverage a search-based approach for optimal plan generation: at each step, we \textit{propose} $K$ probable subsequent actions and \textit{assess} them using a composite of value functions specifically designed for goal-oriented planning. LLMs are employed as both the knowledge base and the assessment tool in the search process, and we illustrate the details in Fig.~\ref{fig:prompt}.}
  \vspace{-5mm}
  \label{fig:framework}
\end{figure}

While the baseline method already demonstrates promising planning results in zero-/few-shot settings, especially for short-term anticipation, its performance diminishes significantly for planning over a long horizon. We hypothesize that deliberate planning mechanisms to explore a combinatorial space of partial solutions and assessment of partial plans with problem-specific heuristic functions are essential to finding the optimal action plan for the procedural tasks~\cite{newell1959report, simon1971human}. 

We propose \model, a search-based framework for optimal plan generation. Figure~\ref{fig:framework} presents an overview of our framework. At each step, we first \textit{propose} $K$ probable subsequent actions using the method described for the baseline in Section~\ref{sec:baseline}. We sample $K$ times from the LLM with the same prompt to capture the uncertainty inherent to the procedural tasks~\cite{zhao2022p3iv, abdelsalam2023gepsan}. Then, we \textit{assess} each proposed action using a composite of value functions specifically tailored for the task of goal-oriented planning. The value functions entail evaluating the coherence and viability of the entire trajectory of predicted actions, leveraging procedural knowledge inherent in the LLM or a task graph derived from few-shot examples. It is noteworthy that this assessment process is crucial to our framework, as it facilitates informed and deliberate decision making~\cite{yao2022react,yao2024tree} and extends beyond the conventional token-level generation process typically used in LLM inference. 
Finally, we employ a breadth-first \textit{search} (BFS) algorithm to identify the optimal action plan according to the assessment scores. A set of the $\tilde{K}$  most promising actions per step is maintained for future evaluation ($\tilde{K} < K$), while the others are pruned to enhance efficiency. Once the best-predicted action at step, $T$ is determined, the optimal action plan can be obtained by backtracking the sequence of predicted actions that led to the generation of $\hat{a}^*_T$. Below, we detail the value functions that contribute to determining the assessment score.

Let us denote the $K$ sampled outputs of the LLM during the $t$-th ``propose'' step as $\left\{s_{t,1}, \dots, s_{t,K} \right\}$, and the projected admissible actions are $\left\{\hat{a}_{t,1}, \dots, \hat{a}_{t,K} \right\}$ (Section~\ref{sec:baseline}). Our assessment score is a weighted sum of four components, each evaluating a different aspect of the action plan:

\begin{figure}[tb]
  \centering
  \includegraphics[width=1.0\textwidth]{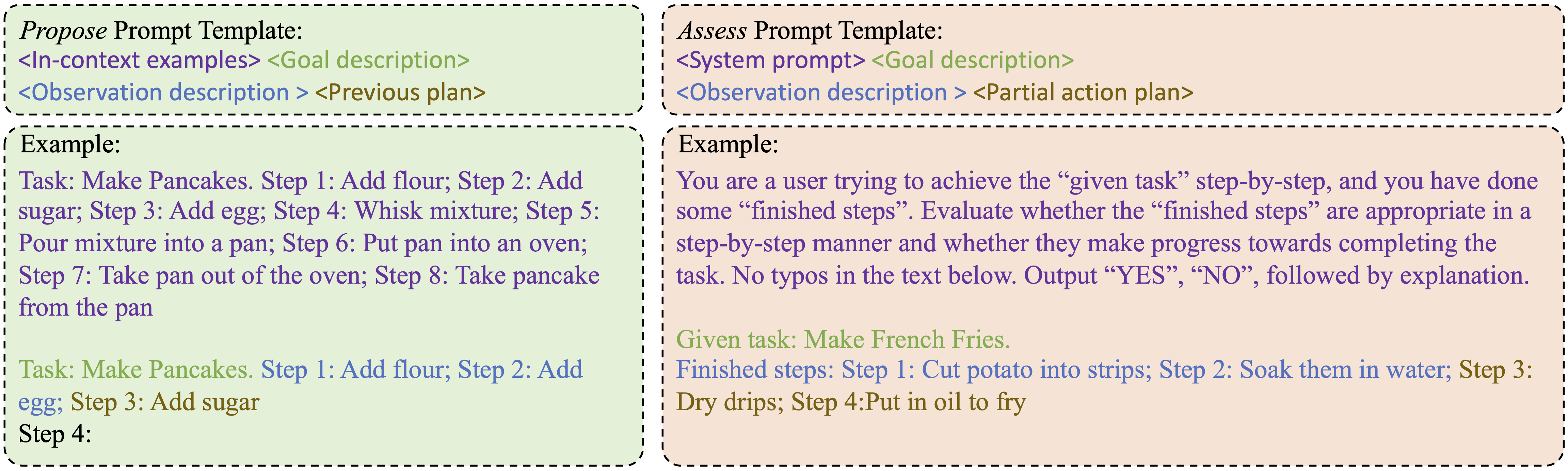}
  \caption{Templates and examples of the LLM prompts we design for subsequent action proposal (left) and partial plan evaluation (right).}
  \vspace{-5mm}
  \label{fig:prompt}
\end{figure}

\vspace{-5mm}
\subsubsection{Text generation score.} This function evaluates the probability of an LLM generating a given description. Intuitively, descriptions of unlikely subsequent actions will yield lower scores. Similar to~\cite{huang2022language}, we compute the mean log probability of the proposed action descriptions as their likelihood scores:
\begin{equation}
     V_G^k = \frac{1}{n_k} \sum_{i=1}^{n_k} \text{log}\  p_\theta(s_{k,i} | \tilde{s}_{k, <i}),
\end{equation}
where $n_k$ is the number of tokens in $s_k$, $\theta$ parameterizes the LLM, and $\tilde{s}_{k, <i}$ denotes the concatenation of the prompt with previously generated tokens.

\vspace{-5mm}
\subsubsection{Text mapping score.} This function assesses the confidence of mapping a free-form description generated by the LLM to an admissible action. Since this mapping is achieved by identifying the admissible action with the highest cosine similarity to the generated description, we use this similarity value as the confidence score:
\begin{equation}
    V_M^k = \max_{\hat{a}_k} \frac{\text{Emb}(s_k)\cdot\text{Emb}(\hat{a}_k)}{\|\text{Emb}(s_k)\|\|\text{Emb}(\hat{a}_k)\|},
\end{equation}
where $\text{Emb}(\cdot)$ is a text embedding function (\eg, Sentence-BERT~\cite{reimers2019sentence}), and $\hat{a}_k$ the admissible action corresponding to highest similarity.

\vspace{-5mm}
\subsubsection{Partial plan evaluation.} This function evaluates the coherence and viability of a partially generated plan up to step $t$. In particular, we repurpose LLMs as the assessment tool by creating a dedicated prompt that requests the LLMs to assess the soundness of the partial plan in order to achieve the given goal (Figure~\ref{fig:prompt}). Unlike the text generation score that emphasizes token-level evaluation, this function evaluates the \textit{semantic} action phrases and takes into account the entire prediction trajectory. This allows the LLM to self-evaluate the progress it makes towards solving the planning problem through a deliberate reasoning process~\cite{yao2024tree,yao2022react}.
To get the evaluation score, we extract the logits of the ``YES'' and ``NO'' tokens from the output of the LLM, followed by softmax normalization:
\begin{equation}
    V_P^k = e^{L_\theta(\texttt{<YES>} | \tilde{p})} / ( e^{L_\theta(\texttt{<YES>} | \tilde{p})} + e^{L_\theta(\texttt{<NO>} | \tilde{p})}),
\end{equation}
Here, $L_\theta$ denotes the logit values generated by the LLM, and $\tilde{p}$ denotes the prompt, including the partially generated plan (Figure~\ref{fig:prompt}).

\vspace{-4mm}
\subsubsection{Few-shot task graph.} This function provides task-specific prior knowledge to our assessment method. In particular, we construct a first-order task graph from the few-shot examples by counting the frequencies of the consecutive action step pairs. Then, we assign a probability value to a partially generated plan $\hat{\mathcal{A}}_t = \left\{\hat{a}_1, \dots, \hat{a}_t \right\}$ using the following equation. Note that we do not use this value function for zero-shot setup.
\begin{equation}
    V_{TG}^k = \prod_{i=1}^{i=t} G_\theta(\hat{a}_{t,i} | \hat{a}_{t, i-1}),
\end{equation}
Here, $G_\theta$ is the few-shot task graph which captures the transitional probabilities of different actions, and $G_\theta(\hat{a}_{t,i} | \hat{a}_{t, i-1})= f(\hat{a}_{i-1}, \hat{a}_{i})/ \sum_{a_y} f(\hat{a}_{i-1}, \hat{a}_{y})$, where $f(\hat{a}_{x}, \hat{a}_{y})$ is the frequencies of the consecutive actions steps $(\hat{a}_{x}, \hat{a}_{y})$ calculated from the few-shot examples.

\subsection{Implementation Details}

We use the Llama-2-70B~\cite{touvron2023llama} as our default Large Language Model. In the visual planning for assistance task, we segment the video history into 1-second clips and utilize VideoCLIP~\cite{xu2021videoclip} to predict the action category of each clip, following prior works~\cite{patel2023pretrained}. Conversely, for the procedural planning task, we adopt the CLIP~\cite{radford2021learning}-based double retrieval model to predict the initial and goal states from the given images, following the previous work~\cite{liu2023language}. In the few-shot setup, we incorporate 3 in-context examples from the training set for the VPA task and 10 in-context examples from the training set for the PP task.

\section{Experimental Setup}
\label{sec:exp_setup}

\subsection{Datasets} 
We conduct our experiments on two widely used instructional video datasets, COIN~\cite{tang2019coin} and CrossTask~\cite{zhukov2019cross}. COIN is a large-scale dataset containing 11,827 videos with 180 different procedural tasks. On the other hand, CrossTask contains 2,750 videos from 18 procedural tasks. We use the standard train/val/test splits and follow the preprocessing steps of prior works to generate samples with input-output pairs for both VPA~\cite{patel2023pretrained} and PP~\cite{chang2020procedure} tasks.

\subsection{Evaluation Protocol} 
\subsubsection{Evaluation Metrics.} Following prior works~\cite{patel2023pretrained, chang2020procedure}, we use three evaluation metrics.
\textbf{Success Rate (SR)} considers an action plan successful only if the predicted action steps and the order match exactly with the ground truth.
\textbf{Mean Accuracy (mAcc)} evaluates the accuracy of individual predicted action steps of a procedural plan compared to their corresponding ground-truth steps at each timestamp.
\textbf{Mean Intersection Over Union (mIoU)} considers predicted and ground-truth action plans as two sets and measures the overlap between them. 

\vspace{-5mm}
\subsubsection{Planning Horizon.} For the VPA task, evaluation is conducted for planning horizons of T=1, 3, and 4, where the model outputs T future steps given the video history and goal. For the PP task, evaluation is performed for planning horizons of T=3 and 4, where the model outputs the start step, end step, and the T-2 intermediate steps given the task name, start and goal images of a task.

\begin{table}[tb]
\centering
\small
\caption{\textbf{Visual Planning for Assistance on COIN.} \model\ achieves the best performance in zero- and few-shot (3 examples) setups. Moreover, our model outperforms the fully supervised state-of-the-art model VLaMP~\cite{patel2023pretrained} by 7.6\%/3.5\%/4.8\% SR for predicting T=1/3/4 future steps using only 3 in-context examples.}
\label{tab:coin vpa}
\vspace{-3mm}
\begin{tabular*}{0.9\textwidth}%
     {@{\extracolsep{\fill}}lccccccc}
\toprule
\multicolumn{1}{c}{\multirow{2}{*}{Method}} & T=1           & \multicolumn{3}{c}{T=3}                       & \multicolumn{3}{c}{T=4}                       \\ \cmidrule(lr){2-2} \cmidrule(lr){3-5} \cmidrule(lr){6-8}
                        & mAcc          & SR            & mAcc          & mIoU          & SR            & mAcc          & mIoU          \\ 
\midrule
\multicolumn{8}{l}{\textbf{Fully Supervised}}  \\
Most Probable           & 0.7           & 1.6           & 4.3           & 6.8           & 1.6           & 8.2           & 15.3          \\
Most Probable w/ goal   & 23.9          & 10.9          & 18.0          & 24.9          & 9.1           & 16.3          & 32.2          \\
DDN~\cite{chang2020procedure}                     & 29.3          & 10.1          & 22.3          & 32.2          & 7.0           & 21.0          & 37.3          \\
VLaMP~\cite{patel2023pretrained}                   & 45.2          & 18.3          & 39.2          & 56.6          & 9.0           & 35.2          & 54.2          \\
\hline
\multicolumn{8}{l}{\textbf{Zero-Shot}}                                                                                                  \\
Random                  & 0.1           & 0.0           & 0.1           & 0.2           & 0.0           & 0.1           & 0.2           \\
Random w /goal          & 24.5          & 1.7           & 21.4          & 42.7          & 0.3           & 20.1          & 47.7          \\
\baseIO~\cite{touvron2023llama} & 28.5 & 2.4 & 25.8 & 43.6 & 0.7 & 26.1 & 45.5 \\
\basePlan~\cite{huang2022language} & 31.1 & 3.1 & 27.1 & 45.5 & 1.1 & 28.5 & 48.0 \\
\model   & 44.5          & 15.3          & 38.0          & 56.9          & 6.1           & 30.6          & 57.0          \\
\hline
\multicolumn{8}{l}{\textbf{Few-Shot}}  \\
\baseIO~\cite{touvron2023llama} & 36.8 & 10.2 & 36.6 & 50.8 & 6.1 & 30.5 & 51.5 \\
\basePlan~\cite{huang2022language} & 38.2 & 11.1 & 40.6 & 52.8 & 6.8 & 33.5 & 53.5\\
\model   & \textbf{52.8} & \textbf{21.8} & \textbf{44.4} & \textbf{64.4} & \textbf{13.8} & \textbf{38.3} & \textbf{66.3} \\
\bottomrule
\end{tabular*}
\vspace{-5mm}
\end{table}
\begin{table}[tb]
\centering
\small
\caption{\textbf{Visual Planning for Assistance on CrossTask.} Our model performs best in zero- and few-shot setups (3 examples). \model\ outperforms the fully supervised state-of-the-art model VLaMP~\cite{patel2023pretrained} by 3.0\% SR while predicting T=4 future steps.}
\label{tab:crosstask vpa}
\vspace{-3mm}
\begin{tabular*}{0.9\textwidth}%
     {@{\extracolsep{\fill}}lccccccc}
\toprule
\multicolumn{1}{c}{\multirow{2}{*}{Method}} & T=1  & \multicolumn{3}{c}{T=3}                       & \multicolumn{3}{c}{T=4}                      \\ \cmidrule(lr){2-2} \cmidrule(lr){3-5} \cmidrule(lr){6-8} 
\multicolumn{1}{c}{}     & mAcc          & SR            & mAcc          & mIOU          & SR           & mAcc          & mIOU          \\ 
\midrule
\multicolumn{8}{l}{\textbf{Fully Supervised}} \\ 
Most Probable                               & 10.4          & 1.7           & 6.1           & 9.9           & 1.3          & 5.5           & 13.9          \\
Most Probable w/ goal                       & 12.4          & 2.4           & 8.9           & 15.5          & 1.5          & 7.9           & 20.5          \\
DDN~\cite{chang2020procedure}                                         & 33.4          & 6.8           & 25.8          & 35.2          & 3.6          & 24.1          & 37.0          \\
LTA~\cite{grauman2022ego4d}                                         & 26.9          & 2.4           & 24.0          & 35.2          & 1.2          & 21.7          & 36.8          \\
VLaMP~\cite{patel2023pretrained}                                       & \textbf{50.6} & 10.3          & 35.3 & 44.0          & 4.4          & 31.7 & 43.4          \\
\hline
\multicolumn{8}{l}{\textbf{Zero-Shot}}  \\
Random                                      & 0.9           & 0.0           & 0.9           & 1.5           & 0.0          & 0.9           & 1.9           \\
Random w /goal                              & 13.2          & 0.3           & 13.4          & 23.6          & 0.0          & 12.7          & 27.8          \\
\baseIO~\cite{touvron2023llama} & 25.8 & 2.1 & 23.2 & 27.7 & 0.4 & 18.9 & 33.0\\
\basePlan~\cite{huang2022language} & 28.7 & 3.1 & 25.6 & 29.9 & 0.8 & 20.0 & 35.6\\
\model                       & 38.7          & 8.7           & 28.5          & 44.1          & 4.6          & 25.8          & 46.8          \\
\hline
\multicolumn{8}{l}{\textbf{Few-Shot}}                                                                                                                      \\
\baseIO~\cite{touvron2023llama} & 31.7 & 4.6 & 29.7 & 35.6 & 1.1 & 22.2 & 41.3 \\
\basePlan~\cite{huang2022language} & 33.1 & 5.8 & 31.3 & 39.6 & 2.1 & 24.7 & 44.2\\
\model                       & 47.8         & \textbf{12.0} & \textbf{36.7}          & \textbf{48.9} & \textbf{7.4} & \textbf{31.9}          & \textbf{51.6} \\ 
\bottomrule
\end{tabular*}
\vspace{-5mm}
\end{table}

\subsection{Baselines}

\textbf{Fully Supervised Baselines.} For the VPA task, we compare with prior state-of-the-art supervised approaches DDN~\cite{chang2020procedure} and VLaMP~\cite{patel2023pretrained}. Following prior works~\cite{patel2023pretrained}, we also compare with a \textit{most probable} baseline, which selects the most probable next action given the previous action, and a \textit{most probable w/ goal} baseline, which chooses the most probable action given the previous action from the actions of a particular task. These two baselines obtain such conditional probabilities by counting frequencies from the training set. On the other hand, for the procedural planning task, we compare against strong supervised methods such as DDN~\cite{chang2020procedure}, PlaTe~\cite{sun2022plate}, Ext-GAIL~\cite{bi2021procedure}, P3IV~\cite{zhao2022p3iv}, E3P~\cite{wang2023event}, and LFP~\cite{liu2023language}.

\vspace{-5mm}
\subsubsection{Zero-Shot Baselines.} We consider four zero-shot baselines that do not use annotated data for the action plan prediction. (1) A \textit{Random} baseline selects an action randomly from the particular dataset at each step. (2) \textit{Random w/ goal} selects an action randomly from the applicable actions to that goal at each step. (3) \textit{\baseIO}~\cite{touvron2023llama}  described in \Cref{sec:baseline}, provides goal and action history to an LLM as input prompts and outputs the future action plan. (4)\textit{\basePlan}~\cite{huang2022language}, an improved version of the \baseIO\, which also generates multiple candidate actions at each step as our model. However, it lacks the deliberate planning mechanism utilizing the propose-assess-based search framework. Both \baseIO\ and \basePlan\ use the same LLM and video history understanding model as \model.

\vspace{-5mm}
\subsubsection{Few-Shot Baselines.} We provide a few in-context examples of action plans from the training set for the \textit{\baseIOshort}~\cite{touvron2023llama} and the \textit{\basePlanshort}~\cite{huang2022language} baselines. The prompt for the VPA is longer than the PP task because additional
action history information is included. Thus, we can
accommodate more in-context examples for the PP task (up to 10) while limiting to 3 examples for the VPA task. 
\section{Results and Analysis}
\label{sec:results}

\begin{table}[tb]
\small
\centering
\caption{\textbf{Procedural Planning on COIN.} \model\ achieves the best performance in zero-shot and few-shot (10 examples) settings. Compared to the fully supervised state-of-the-art model LFP~\cite{liu2023language}, our model achieves significantly better performance (+4.81\% SR) for longer horizon T=4 steps.}
\label{tab:coin pp}
\vspace{-3mm}
\begin{tabular*}{0.9\textwidth}%
     {@{\extracolsep{\fill}}lcccccc}
\toprule
\multirow{2}{*}{Method}  & \multicolumn{3}{c}{T=3}                          & \multicolumn{3}{c}{T=4}  \\ 
\cmidrule(lr){2-4} \cmidrule(lr){5-7}  & SR             & mAcc           & mIOU           & SR             & mAcc           & mIOU    \\ 
\midrule
\multicolumn{7}{l}{\textbf{Fully Supervised}}                                                                                  \\
DDN~\cite{chang2020procedure}                      & 13.90          & 20.19          & 64.78          & 11.13          & 17.71          & 68.06          \\
P3IV~\cite{zhao2022p3iv}                     & 15.40          & 21.67          & 76.31          & 11.32          & 18.85          & 70.53          \\
E3P~\cite{wang2023event}                      & 19.57          & 31.42          & 84.95          & 13.59          & 26.72          & \textbf{84.72} \\
PPDP~\cite{wang2023pdpp}                     & 21.33          & 45.62          & 51.82          & 14.41          & 44.10          & 51.39          \\
LFP~\cite{liu2023language}                      & \textbf{30.64} & 54.72          & 76.86          & 15.97          & \textbf{50.70} & 75.30          \\ \hline
\multicolumn{7}{l}{\textbf{Zero-Shot}}                                                                                         \\
Random                   & 0.01           & 0.01           & 2.47           & 0.01           & 0.01           & 2.32           \\
\baseIO~\cite{touvron2023llama}   & 13.04          & 45.15          & 64.45          & 4.46           & 38.55          & 73.96          \\
\basePlan~\cite{huang2022language} & 14.15          & 47.07          & 67.41          & 5.46           & 40.18          & 75.17          \\
\model    & 18.44          & 50.63          & 75.64          & 9.07           & 42.72          & 80.83          \\ \hline
\multicolumn{7}{l}{\textbf{Few-Shot}}                                                                                          \\
\baseIO   & 21.79          & 51.17          & 68.17          & 11.25          & 42.17          & 76.17          \\
\basePlan~\cite{huang2022language}   & 23.65          & 52.55          & 69.00          & 12.20          & 43.19          & 77.71          \\
\model    & 29.20          & \textbf{54.76} & \textbf{78.02} & \textbf{20.78} & 49.07          & 78.93          \\ 
\bottomrule
\end{tabular*}
\vspace{-5mm}
\end{table}
\begin{table}[tb]
\centering
\small
\caption{\textbf{Procedural Planning in CrossTask dataset.} Our model achieves the best performance in both zero- and few-shot setups. It also achieves comparable performance as the state-of-the-art fully supervised models despite using only 10 labeled examples.}
\label{tab:crosstask pp}
\vspace{-3mm}
\begin{tabular*}{0.9\textwidth}%
     {@{\extracolsep{\fill}}lcccccc}
\toprule
\multirow{2}{*}{Method}  & \multicolumn{3}{c}{T=3}                          & \multicolumn{3}{c}{T=4}  \\ 
\cmidrule(lr){2-4} \cmidrule(lr){5-7}  & SR             & mAcc           & mIOU           & SR             & mAcc           & mIOU    \\ 
\midrule
\multicolumn{7}{l}{\textbf{Fully Supervised}}                                                                                  \\
DDN~\cite{chang2020procedure}                      & 12.18          & 31.29          & 47.48          & 5.97           & 27.10          & 48.46          \\
PlaTe~\cite{sun2022plate}                    & 16.00          & 36.17          & 65.91          & 14.00          & 35.29          & 55.36          \\
Ext-GAIL~\cite{bi2021procedure}                 & 21.27          & 49.46          & 73.89          & 13.40          & 44.16          & 70.01          \\
P3IV~\cite{zhao2022p3iv}                     & 23.34          & 49.96          & 73.89          & 13.40          & 44.16          & 70.01          \\
E3P~\cite{wang2023event}                      & 26.40          & 53.02          & 74.05          & \textbf{16.49} & 48.00          & 70.16          \\
PPDP~\cite{wang2023pdpp}                     & 26.47          & 55.35          & 58.95          & 15.40          & 49.42          & 56.99          \\
LFP                      & \textbf{30.55} & \textbf{59.59} & \textbf{76.86} & 15.97          & 50.70          & \textbf{75.30} \\
\hline
\multicolumn{7}{l}{\textbf{Zero-Shot}}                                                                                         \\
Random                   & 0.01           & 0.94           & 1.66           & 0.01           & 1.83           & 1.66           \\
\baseIO~\cite{touvron2023llama}   & 9.04           & 44.30          & 61.16          & 3.27           & 36.61          & 64.91          \\
\basePlan~\cite{huang2022language} & 11.21          & 47.07          & 65.98          & 5.11           & 38.18          & 68.66          \\
\model    & 14.60          & 52.60          & 68.38          & 9.89           & 40.85          & 70.35          \\
\hline
\multicolumn{7}{l}{\textbf{Few-Shot}}  \\
\baseIO   & 20.79          & 53.25          & 69.96          & 8.97           & 44.32          & 68.81          \\
\basePlan & 22.67          & 54.48          & 70.10          & 11.06          & 46.63          & 70.14          \\
\model    & 28.85          & 58.12          & 75.36          & 15.45          & \textbf{51.51} & 72.61 \\
\bottomrule
\end{tabular*}
\vspace{-5mm}
\end{table}

\subsection{Main Results on Visual Planning for Assistance}
\label{sec:results vpa}

We conduct experiments on both the COIN~\cite{tang2019coin} and CrossTask~\cite{zhukov2019cross} datasets for the VPA task, with results presented in \Cref{tab:coin vpa} and \Cref{tab:crosstask vpa}. Our analysis reveals several noteworthy findings. Firstly, our proposed \model\ model demonstrates superior performance across all metrics in both zero-shot and few-shot setups. In the zero-shot scenario, \model\ surpasses the \baseIO, by 12.9\%/5.4\% SR on COIN and 6.6\%/4.4\% SR on CrossTask for predicting T=3/4 future steps. In the few-shot setting, \model\ outperforms \baseIO\ by 11.6\%/7.7\%  SR on COIN and 7.4\%/6.3\% SR on CrossTask for T=3/4. This demonstrates the inadequacy of the simple LLM prompt-based techniques for this task, and highlights the effectiveness of the deliberate planning mechanism of our model. Secondly, our model consistently outperforms the \basePlanshort\ baseline by a considerable margin across all metrics in both datasets. Specifically, \model\ achieves a 12.2\% higher SR on COIN and 5.6\% higher SR on CrossTask in the zero-shot setting for predicting T=3 future steps. This shows the significance of the search-based planning mechanism of our model. Lastly, \model\ outperforms the state-of-the-art fully supervised method VLaMP~\cite{patel2023pretrained} using only 3 examples by 3.5\%/4.8\% SR on COIN and 1.7\%/3.0\% SR on CrossTask for T=3/4 steps. This highlights the superior planning ability of our framework for the VPA task, even with limited annotated samples. 

\subsection{Main Results on Procedural Planning}
\label{sec:results pp}

We conduct experiments on both the COIN~\cite{tang2019coin} and CrossTask~\cite{zhukov2019cross} datasets for the PP task and show the results in \Cref{tab:coin pp} and \Cref{tab:crosstask pp}. Similar to the VPA task, our model performs strongly in both zero-shot and few-shot scenarios. Specifically, in the zero-shot setting, \model\ surpasses the standard LLM baseline, \baseIO, by 5.40\%/7.41\% SR in COIN and 5.56\%/6.62\% SR in CrossTask datasets for action horizon T=3/4. This highlights the efficacy of our search-based deliberate planning mechanism compared to the standard LLM baseline. Furthermore, our model outperforms the \basePlan\ baseline by 4.29\% SR in COIN and 3.39\% SR in CrossTask for T=3 steps in zero-shot setups, demonstrating the effectiveness of our explicit planning technique. Finally, compared to the state-of-the-art supervised model LFP, our few-shot (10 examples) \model\ achieves comparable performances, even outperforming in several cases (e.g., +4.81\% boost in T=4 steps for the COIN dataset).

\subsection{Ablation Studies}
\label{sec:ablation}

\begin{table}[tb]
\centering
\small
\caption{\textbf{Importance of Value Functions.} Combining all four value functions leads to the best performance in both VPA and PP tasks.}
\label{tab:ablation value}
\vspace{-3mm}
\begin{tabular*}{1\textwidth}%
     {@{\extracolsep{\fill}}ccccccccc}
\toprule
\multicolumn{1}{c}{\multirow{2}{*}{\begin{tabular}[c]{@{}c@{}}Generation\\ Score\end{tabular}}} & \multicolumn{1}{c}{\multirow{2}{*}{\begin{tabular}[c]{@{}c@{}}Mapping \\ Score\end{tabular}}} & \multicolumn{1}{c}{\multirow{2}{*}{\begin{tabular}[c]{@{}c@{}}Task \\ Graph\end{tabular}}} & \multicolumn{1}{c}{\multirow{2}{*}{\begin{tabular}[c]{@{}c@{}}Partial\\ Plan\end{tabular}}} & \multicolumn{3}{c}{VPA} & \multicolumn{2}{c}{PP}            \\ \cmidrule(lr){5-7} \cmidrule(lr){8-9}  
\multicolumn{1}{c}{} & \multicolumn{1}{c}{} & \multicolumn{1}{c}{} & \multicolumn{1}{c}{}                                                                 & T=1    & T=3    & T=4   & T=3                       & T=4   \\ 
\midrule
\cmark & & & & 40.18  & 11.60  & 6.70   & 18.25      & 11.33     \\
& \cmark & & & 39.67  & 10.21  & 6.10   & 17.25      & 11.30     \\
& & \cmark & &  38.23  & 9.35   & 5.50   & 19.57      & 13.19     \\
& & & \cmark &  47.10  & 16.10   & 9.10   & 24.44      & 16.20     \\
\cmark  & \cmark  & \cmark  &  & 45.66  & 16.96  & 9.51  & 24.38 & 16.89 \\
\cmark & \cmark  &  & \cmark & 50.67  & 19.91  & 12.33 & 27.81 & 19.03 \\
\cmark  &  & \cmark  & \cmark  & 49.69  & 19.71  & 12.01 & 26.76 & 18.67 \\
\cmark  & \cmark  & \cmark  & \cmark & \textbf{52.20}  & \textbf{21.08}  & \textbf{13.80} & \textbf{29.69} & \textbf{20.78}\\
\bottomrule
\end{tabular*}
\vspace{-2mm}
\end{table}

\subsubsection{Importance of Value Functions.}

In this section, we conduct an ablation study to evaluate the significance of four value functions in our goal-oriented planning framework for instructional videos: text generation score, text mapping score, partial plan evaluation, and few-shot task graph (detailed explanations can be found in~\Cref{sec:model}). The performance of different value functions and their combinations is presented in \Cref{tab:ablation value}. First, we observe that the LLM-based partial plan evaluation function achieves the highest success rate when using only one value function as the assessment method (rows 1-4). Additionally, when using a combination of three value functions, the removal of the partial plan evaluation function results in the largest drop in performance (-4.12\%/5.31\% SR in VPA/PP for T=3 steps) (rows 5-9). This highlights the efficacy of utilizing LLM as the assessment tool for evaluating action plans in goal-oriented tasks. Moreover, we note that the combination of text generation score, text mapping score, and partial plan evaluation yields the second-highest performance across all metrics. Finally, the combination of all four value functions leads to the highest overall performance, demonstrating the importance of each proposed value function for goal-oriented planning tasks.

\begin{table}[tb]
\centering
\small
\caption{\textbf{Performance with Ground-Truth Action History.} Ground-truth action steps yields significantly better performance than predicted actions on COIN, indicating that enhancing the visual models could improve the performance of \model.}
\vspace{-3mm}
\label{tab:ablation gt}
\begin{tabular*}{0.7\textwidth}%
     {@{\extracolsep{\fill}}ccccccc}
\toprule
\multirow{2}{*}{\begin{tabular}[c]{@{}c@{}}Plan Horizon\\ (T)\end{tabular}} & \multicolumn{3}{c}{VPA} & \multicolumn{3}{c}{PP} \\ \cmidrule(lr){2-4} \cmidrule(lr){5-7}
& SR     & mAcc   & mIOU  & SR     & mAcc  & mIOU  \\
\midrule
1                                       & 65.96  & 65.96  & 65.96  & -      & -     & -     \\
3                                       & 41.33  & 57.16  & 75.99  & 66.13  & 66.13 & 66.13 \\
4                                       & 25.69  & 44.16  & 70.10  & 38.19  & 54.17 & 63.31 \\ \bottomrule
\end{tabular*}
\vspace{-5mm}
\end{table}


\vspace{-5mm}
\subsubsection{Performance with Ground-Truth Action History.} In this section, we investigate the performance of our model using ground-truth action steps instead of predicted action steps of visual observations. Specifically, we use ground-truth action history and the goal as input prompts for the LLM in the VPA task. On the other hand, for the PP task, we employ ground-truth start and end action steps as input prompts for the LLM. The results in \Cref{tab:ablation gt} indicate that ground-truth observations yield significantly better performance than predicted observations. For the VPA task, \model\ achieves improvements of 13.16\%, 19.53\%, and 11.89\% in SR for T=1, 2, and 3 steps, respectively, when using ground-truth action history instead of predicted action history. On the other hand, in the PP task, the SR improves by 36.93\% and 17.41\% for T=3 and 4 steps by using ground-truth start and goal steps instead of predicted steps. These findings suggest that improving the visual understanding model can further enhance the performance of our framework.

\begin{table}[tb]
\centering
\small
\caption{\textbf{Ablation of LLM Sizes.} Utilizing stronger LLMs results in better performance. We show the success rate for different planning horizons (T) for both tasks.}
\vspace{-2mm}
\label{tab:ablation model size}
\begin{tabular*}{0.7\textwidth}%
     {@{\extracolsep{\fill}}ccccccc}
\toprule
\multirow{2}{*}{LLM Size} & \multicolumn{3}{c}{VPA} & \multicolumn{2}{c}{PP} \\ \cmidrule(lr){2-4} \cmidrule(lr){5-6} & T=1       & T=3      & T=4     & T=3          & T=4         \\
\midrule
Llama-2-7B                  & 48.60  & 19.61  & 11.71 & 26.78      & 17.72     \\
Llama-2-30B                 & 50.16  & 20.07  & 12.28 & 28.10      & 18.96     \\
Llama-2-70B                 & \textbf{52.20}  & \textbf{21.08}  & \textbf{13.80} & \textbf{29.69 }     & \textbf{20.78}    \\ 
\bottomrule
\end{tabular*}
\vspace{-2mm}
\end{table}

\vspace{-5mm}
\subsubsection{Ablation of LLM Sizes.} \Cref{tab:ablation model size} presents the success rate of our few-shot model on the COIN dataset~\cite{tang2019coin} using different LLMs. Firstly, we note that even the smallest Llama-2-7B LLM, \model\ demonstrates good performance. For instance, it surpasses the fully supervised state-of-the-art model VLaMP~\cite{patel2023pretrained} by 1.31\% and 2.71\% SR for T=3 and 4 in the VPA task while utilizing only 3 examples. Secondly, we observe a consistent improvement in performance with increasing LLM sizes. This indicates that leveraging more powerful LLMs has the potential to further improve the performance of the \model\ framework.

\begin{table}[tb]
\centering
\small
\caption{\textbf{Ablation of In-context Examples.} Performance increases with the increase of in-context examples. We could fit upto 3 examples for the VPA task and 10 examples for the PP task in the LLM prompt.}
\vspace{-2mm}
\label{tab:ablation example shots}
\begin{tabular*}{0.7\textwidth}%
     {@{\extracolsep{\fill}}ccccccc}
\toprule
\multirow{2}{*}{Example Shots} & \multicolumn{3}{c}{VPA} & \multicolumn{2}{c}{PP} \\ \cmidrule(lr){2-4} \cmidrule(lr){5-6} & T=1       & T=3      & T=4     & T=3          & T=4         \\
\midrule
0  & 44.50  & 15.30  & 6.10  & 18.44      & 9.07      \\
1  & 48.50  & 18.17  & 10.38 & 21.90      & 14.13     \\
3  & \textbf{52.20}  & \textbf{21.08}  & \textbf{13.80} & 25.25      & 17.67     \\
10 & -      & -      & -     & \textbf{29.69}      & \textbf{20.78}     \\
\bottomrule
\end{tabular*}
\vspace{-4mm}
\end{table}


\vspace{-4mm}
\subsubsection{Ablation of In-context Examples.} Finally, we present the performance (SR) of our model with varying numbers of in-context examples in \Cref{tab:ablation example shots}. We observe a consistent increase in performance with the increase in the number of training examples. This suggests that while our zero/few-shot framework demonstrates strong performance in goal-oriented planning tasks, utilizing more annotated training data can lead to even better results.
\section{Conclusions}
\label{sec:conclusion}

We introduced \model, a goal-oriented planning framework designed for challenging zero/few-shot learning settings in instructional videos. Our framework integrates LLMs and search-based techniques for superior long-horizon planning, achieving competitive or even better results than fully supervised approaches. In the future, we will extend our framework to even more complex tasks and video inputs that may last several hours. Furthermore, we will explore the design of more sophisticated search algorithms and value functions for improved zero-shot performance of our system. 

\noindent\textbf{Acknowledgements.} We thank Yan-Bo Lin, Feng Cheng, Ce Zhang, Yue Yang, and Soumitri Chattopadhyay for their helpful discussions. This work was supported by the Sony Faculty Innovation Award, Laboratory for Analytic Sciences via NC State University, ONR Award N00014-23-1-2356.

\setcounter{section}{0}
\setcounter{figure}{0}
\setcounter{table}{0}

\renewcommand{\thesection}{S\arabic{section}}
\renewcommand{\thetable}{S\arabic{table}}
\renewcommand{\thefigure}{S\arabic{figure}}

\begin{center} 
\Large{\textbf{Supplementary Material}}
\end{center} 

Our supplementary material includes additional related works, additional implementation details, and qualitative results. We will release the code upon acceptance of the paper.

\section{Additional Related Works}

Several recent studies have introduced various approaches for future action anticipation from videos~\cite{sener2019zero, gao2023assistgpt}. Additionally, several intriguing methodologies have emerged for procedural planning in instructional videos. Notable among these are prompt-based techniques, including commonsense prompting~\cite{lu2022neuro} and multimodal image-text prompting~\cite{lu2023multimodal}. Another work~\cite{li2023skip} proposed a condensed action space learning method for procedural planning. In contrast, our work presents a unified framework for different goal-oriented planning tasks (i.e., visual planning for assistance and procedural planning), employing a deliberate propose, assess, and search technique.

Concurrently, there is a growing body of literature exploring reasoning and planning with large language models (LLMs). Various studies leverage LLMs with different tools to tackle complex tasks~\cite{schick2024toolformer, gao2023assistgpt, sun2023pearl}. Moreover, there has been significant research on planning in multimodal domains using LLMs~\cite{gao2023assistgpt, rose2023visual, du2023video}. On the contrary, our approach harnesses LLMs as both a knowledge base and an assessment tool for goal-oriented planning in instructional videos.

\section{Additional Implementation Details}

\noindent\textbf{Video History Understanding Model for VPA task.} We use VideoCLIP~\cite{xu2021videoclip} as our video history understanding model following the prior work VLaMP~\cite{patel2023pretrained} to ensure fair comparison. To obtain action steps prediction from the video history, we divide the video into fixed-length windows of 1-second clips. Then, we classify each clip using the pretrained VideoCLIP~\cite{xu2021videoclip} model. Afterward, we marge the consecutive clips with the same action category as the one step prediction, and thus, we get a sequence of action history from the video. 

\vspace{3mm}
\noindent\textbf{Step Classification Model for PP task.} Following the prior work~\cite{liu2023language}, we use a retrieval model to predict the initial and goal action steps from the visual observations. In particular, we utilize a BLIP~\cite{li2022blip} base model to retrieve the initial and goal steps simultaneously from the provided images. We follow the double retrieval method proposed by~\cite{liu2023language} and finetune a BLIP~\cite{li2022blip}-base model following their approach. Please refer to~\cite{liu2023language} for more details on the retrieval-based step classification for this task. 

\vspace{3mm}
\noindent\textbf{Combining Value Functions.} As described in Section 3.3 of the main paper, we utilize four value functions (text generation score, text mapping score, task graph score, and partial plan evaluation) to guide the breadth-first search of \model\ to find the optimal action plan. In particular, we utilize a weighted combination of all four functions as the assessment criteria for generated partial plans at each step. We find the optimal weights of each value function from the held-out validation set and use them for the test set. Specifically, we use the following function for the visual planning for assistance task.

\begin{equation}
    V^k = 0.2*V^k_G + 0.1*V^k_M + 0.1*V^k_{TG} + 0.7*V^k_P
\end{equation}

Here, $V^k_G$ is the text generation score of a particular sample, $V^k_M$ is the text mapping score, $V^k_{TG}$ is the task graph score, $V^k_P$ is the partial plan evaluation score, and $V^k$ is the combined value score. On the other hand, we use the following function for the procedural planning task.

\begin{equation}
    V^k = 0.1*V^k_G + 0.1*V^k_M + 0.3*V^k_{TG} + 0.5*V^k_P
\end{equation}

\section{Qualitative Results}

We present qualitative results on both visual planning for assistance (VPA) and procedural planning (PP) tasks. 

\begin{figure}[t]
  \centering
  \includegraphics[width=1\textwidth]{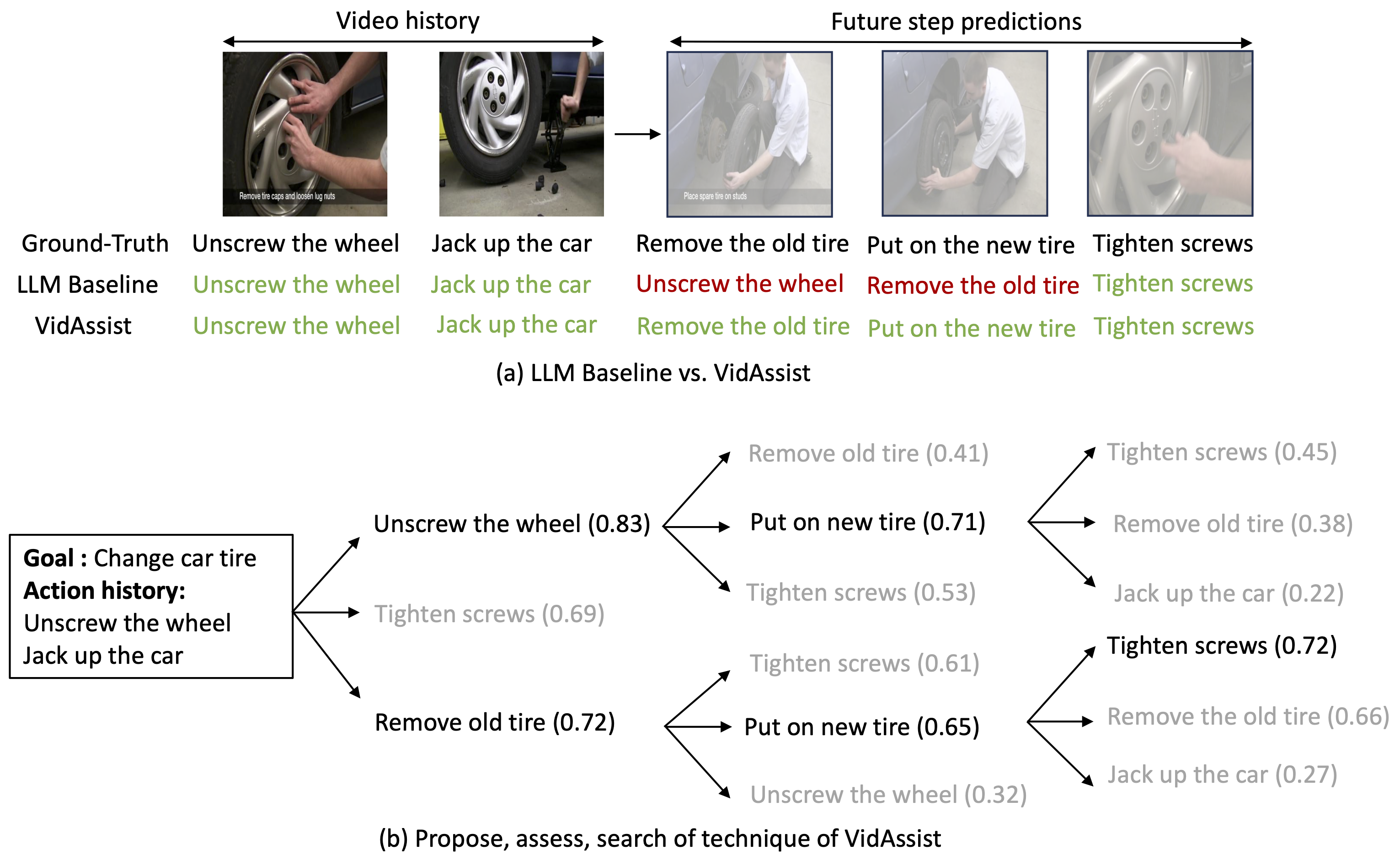}
  \caption{Example of visual planning for assistance in COIN dataset. (a) \model\ successfully predicts the future action steps while the LLM baseline fails. (b) Visualization of the proposed search technique with intermediate steps and value scores. We only show three generated actions at each step for brevity and clarity.}
  \label{fig:qualitative_vpa_coin}
\end{figure}

\begin{figure}[t]
  \centering
  \includegraphics[width=1\textwidth]{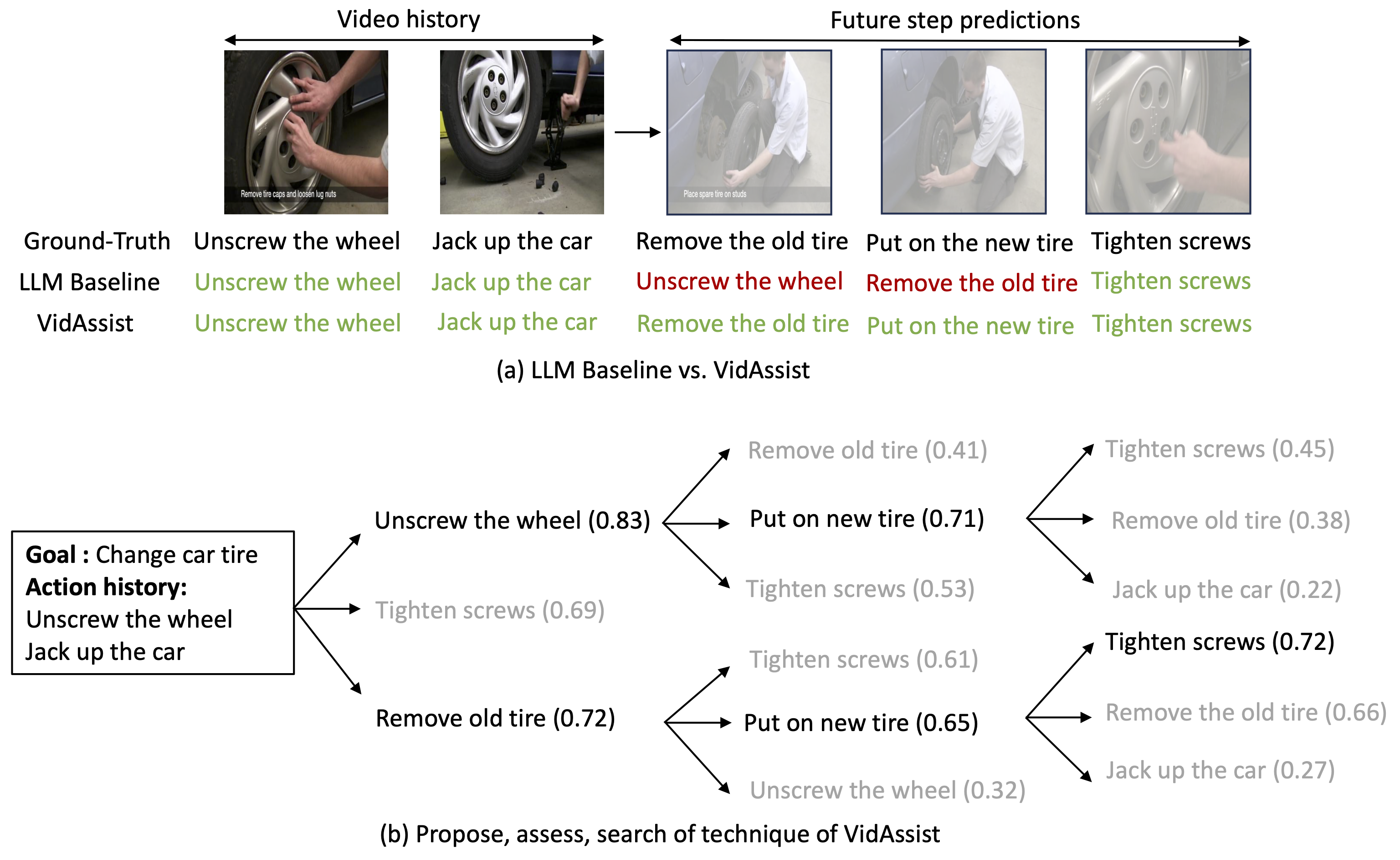}
  \caption{\textbf{Example of visual planning for assistance in CrossTask dataset.} (a) \model\ successfully predicts the future action steps while the LLM baseline fails. (b) Visualization of the proposed search technique with intermediate steps and value scores. We only show three generated actions at each step for brevity and clarity.}
  \label{fig:qualitative_vpa_ct}
\end{figure}

\subsection{Qualitative Results on VPA task} We show one example from the COIN dataset in \Cref{fig:qualitative_vpa_coin} and one example from the CrossTask dataset in \Cref{fig:qualitative_vpa_ct}. In both examples, we observe that while our model successfully predicts the future action plan, the LLM baseline fails to predict the correct action steps. Moreover, we show the propose, assess and search technique of the \model\ model in \Cref{fig:qualitative_vpa_coin} (b) and \Cref{fig:qualitative_vpa_ct} (b). We notice that our model is able to search the optimal action plan from the generated trees utilizing the proposed value scores. For instance, in \Cref{fig:qualitative_vpa_coin} (b), we observe that the model assigns the highest score to the `Unscrew the wheel' action in step 1; however, the correct action for step 1 in `remove old tire'. Nevertheless, the \model\ rectifies its error in the final step by choosing the highest value path at step 3. Thus, it finds the optimal plan for the particular task. This shows the effectiveness of our search-based technique and the proposed value functions for this particular task.

\begin{figure}[t]
  \centering
  \includegraphics[width=1\textwidth]{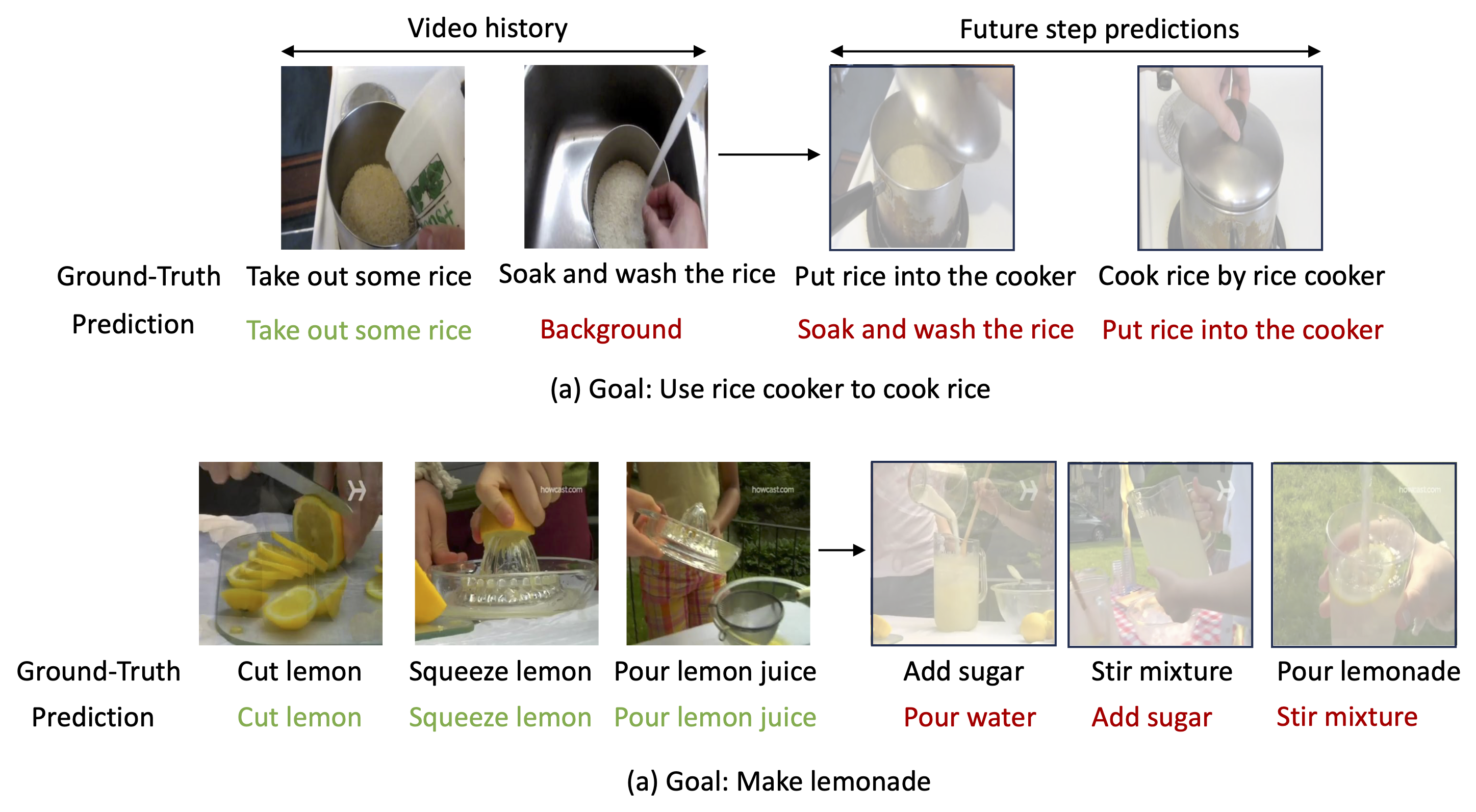}
  \caption{\textbf{Examples failure cases of \model\ in visual planning for assistance.} (a) The video history understanding model fails to identify the correct history steps, which leads to the wrong future step predictions. However, our model makes valid predictions based on the predicted action history. (b) \model\ makes reasonable predictions, though they do not match with the ground truth.}
  \label{fig:qualitative_vpa_failure}
\end{figure}

We also show two failure cases of our model in \Cref{fig:qualitative_vpa_failure}. In \Cref{fig:qualitative_vpa_failure} (a), we observe that the video history prediction model fails to identify the correct steps. From the provided video history, only one step, `take out some rice,' was identified, while the user also performed the step `soak and wash the rice` in the video. Thus, our model predicts `soak and wash the rice` and `put rice into the cooker` as future steps. This shows though the error is coming from the video history understanding model, \model\ still makes valid predictions based on the observed history. This also indicates that the performance of our model can be further enhanced by improving the video history understanding model. On the other hand, in \Cref{fig:qualitative_vpa_failure} (b), we observe that although the predicted future action steps do not match perfectly with the ground-truth actions, our model still makes reasonable predictions for the particular task. 

\begin{figure}[t]
  \centering
  \includegraphics[width=1\textwidth]{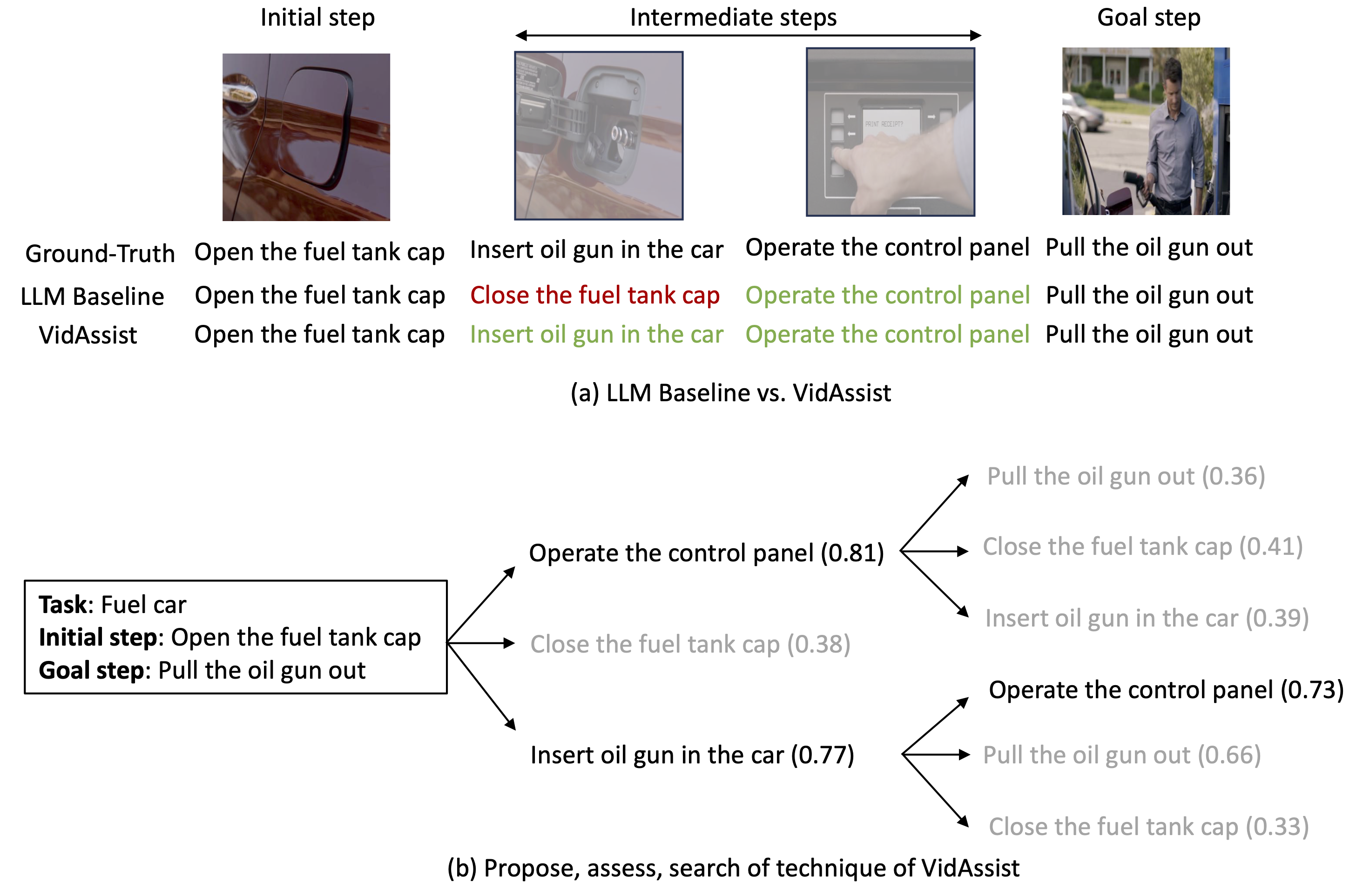}
  \caption{\textbf{Example of procedural planning for assistance in COIN dataset.} (a) \model\ successfully predicts the future action steps while the LLM baseline fails. (b) Visualization of the proposed search technique with intermediate steps and value scores. We only show three generated actions at each step for brevity and clarity.}
  \label{fig:qualitative_pp_coin}
\end{figure}

\begin{figure}[t]
  \centering
  \includegraphics[width=0.8\textwidth]{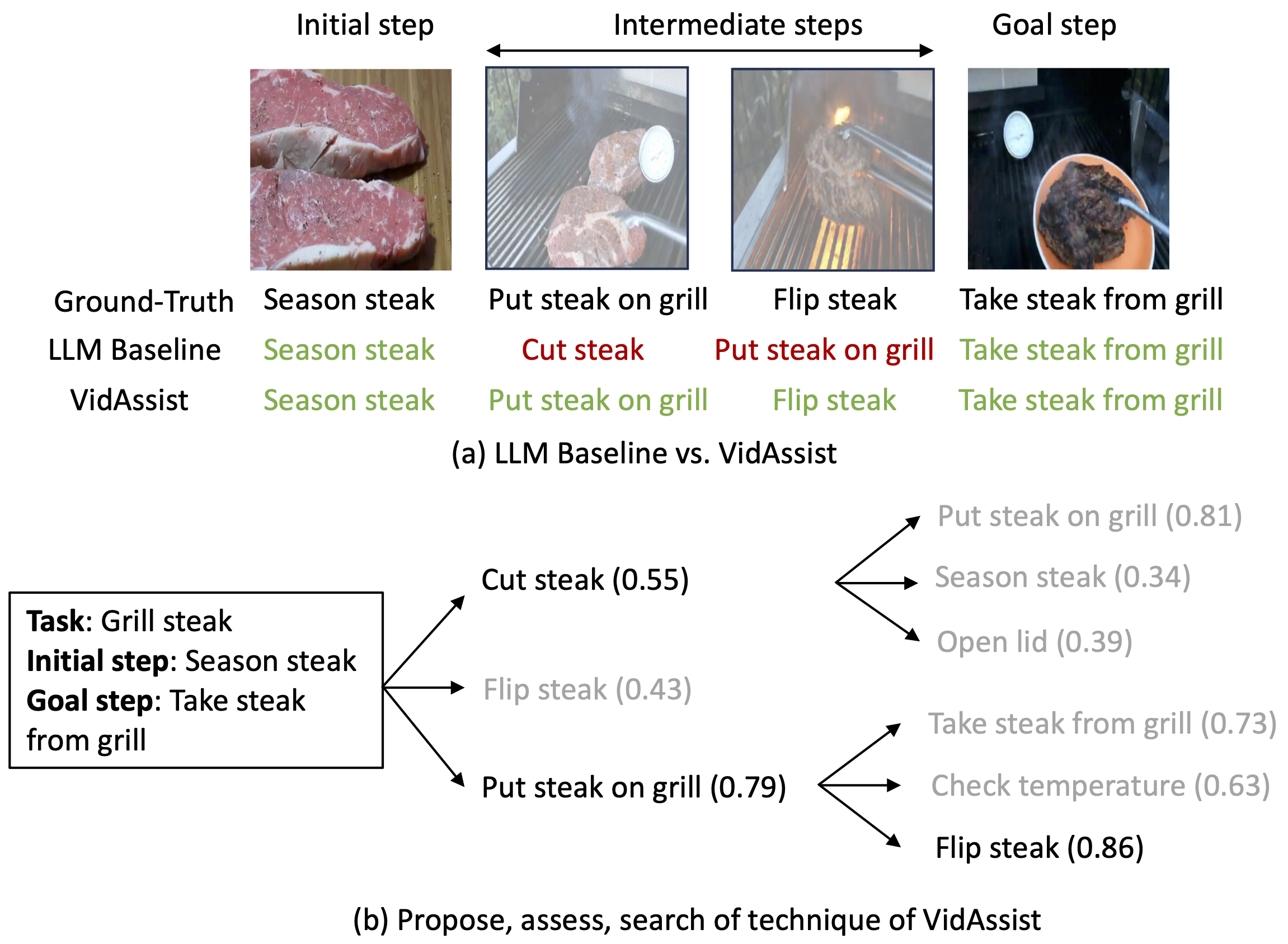}
  \caption{\small{\textbf{Example of procedural planning for assistance in CrossTask dataset.} (a) \model\ successfully predicts the future action steps while the LLM baseline fails. (b) Visualization of the proposed search technique with intermediate steps and value scores. We only show three generated actions at each step.}}
  \label{fig:qualitative_pp_ct}
\end{figure}

\subsection{Qualitative Results on PP task} For the procedural planning task, We show one example from the COIN dataset in \Cref{fig:qualitative_pp_coin} and one example from the CrossTask dataset in \Cref{fig:qualitative_pp_ct}. \model\ successfully predicts the correct future action plan in both cases while the LLM baseline fails. We also visualize the propose, assess and search technique of the \model\ model in \Cref{fig:qualitative_pp_coin} (b) and \Cref{fig:qualitative_pp_ct} (b). We observe that our model is able to search the optimal action plan from the generated trees utilizing the proposed value scores. This demonstrates the effectiveness of our search-based technique and the proposed value functions for the procedural planning task.

\begin{figure}[t]
  \centering
  \includegraphics[width=0.8\textwidth]{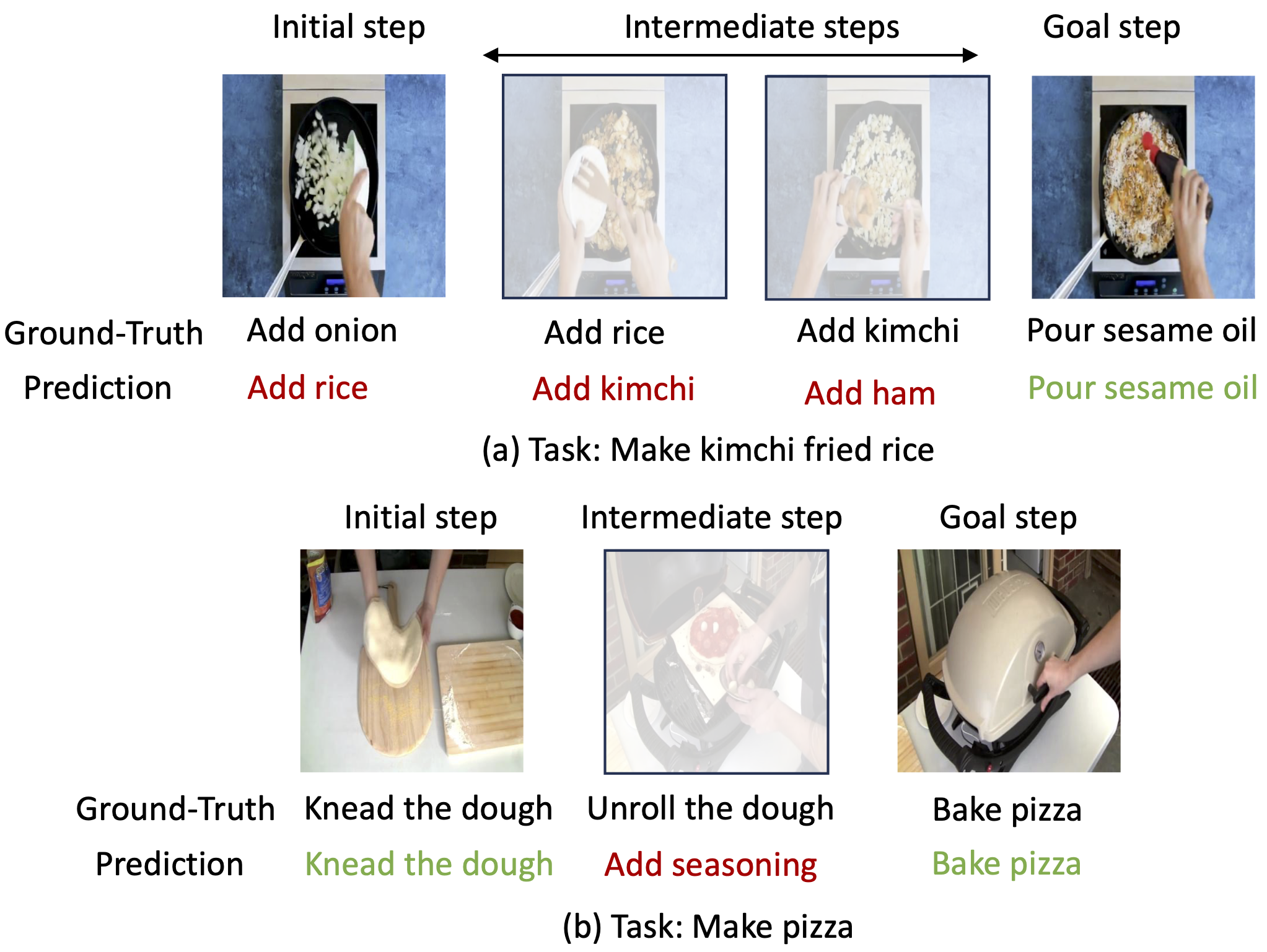}
  \caption{\textbf{Examples failure cases of \model\ in procdural planning.} (a) The step classification model fails to identify the correct initial steps, which leads to the wrong intermediate step predictions. However, our model makes valid predictions based on the predicted initial and goal steps. (b) \model\ makes a reasonable prediction for the intermediate step, though it does not match with the ground truth.}
  \label{fig:qualitative_pp_failure}
\end{figure}

Moreover, we present two failure cases of our model in \Cref{fig:qualitative_pp_failure}. In \Cref{fig:qualitative_vpa_failure} (a), we observe that the step classification model fails to predict the correct initial step. However, the intermediate steps generated by our model are reasonable considering the predicted initial and goal steps. Therefore, the error stems from the step classification model in this case rather than our LLM-based search technique. This also indicates that the performance of our model can be further enhanced by improving the video step classification model. On the other hand, in \Cref{fig:qualitative_vpa_failure} (b), we observe that although the predicted intermediate step does not match perfectly with the ground truth, our model still makes a valid prediction for the particular task.

\clearpage

%
%
\bibliographystyle{splncs04}
\bibliography{main}
\end{document}